\documentclass[10pt,twocolumn,letterpaper]{article}

\usepackage[pagenumbers]{cvpr} 

\definecolor{cvprblue}{rgb}{0.21,0.49,0.74}
\usepackage[pagebackref,breaklinks,colorlinks,allcolors=cvprblue]{hyperref}

\usepackage{booktabs, xcolor, colortbl}
\usepackage{graphicx}
\usepackage{array}
\usepackage{amsmath}
\usepackage{siunitx}
\definecolor{darkgreen}{RGB}{0,175,0}
\newcommand{\increase}[1]{\textcolor{darkgreen}{\textuparrow{}}}
\newcommand{\decrease}[1]{\textcolor{red}{\textdownarrow{}}}
\usepackage{multirow}
\usepackage{algorithm,algorithmicx,algpseudocode}

\def\paperID{*****} 
\def\confName{CVPR}
\def\confYear{2025}

\title{A Probabilistic Jump-Diffusion Framework for Open-World Egocentric Activity Recognition}

\author{Sanjoy Kundu, Shanmukha Vellamcheti, Sathyanarayanan N. Aakur\\
CSSE Department, Auburn University\\
Auburn, Alabama, USA 36849\\
{\tt\small \{szk0266, szv0080, san0028\}@auburn.edu}
}
\begin{document}
\maketitle
\begin{abstract}
Open-world egocentric activity recognition poses a fundamental challenge due to its unconstrained nature, requiring models to infer unseen activities from an expansive, partially observed search space. We introduce ProbRes, a Probabilistic Residual search framework based on jump-diffusion that efficiently navigates this space by balancing prior-guided exploration with likelihood-driven exploitation. Our approach integrates structured commonsense priors to construct a semantically coherent search space, adaptively refines predictions using Vision-Language Models (VLMs) and employs a stochastic search mechanism to locate high-likelihood activity labels while minimizing exhaustive enumeration efficiently. We systematically evaluate ProbRes across multiple openness levels (L0–L3), demonstrating its adaptability to increasing search space complexity. In addition to achieving state-of-the-art performance on benchmark datasets (GTEA Gaze, GTEA Gaze+, EPIC-Kitchens, and Charades-Ego), we establish a clear taxonomy for open-world recognition, delineating the challenges and methodological advancements necessary for egocentric activity understanding. 

\end{abstract}    

\section{Introduction}

\begin{figure}[t]
    \centering
    \includegraphics[width=0.99\columnwidth]{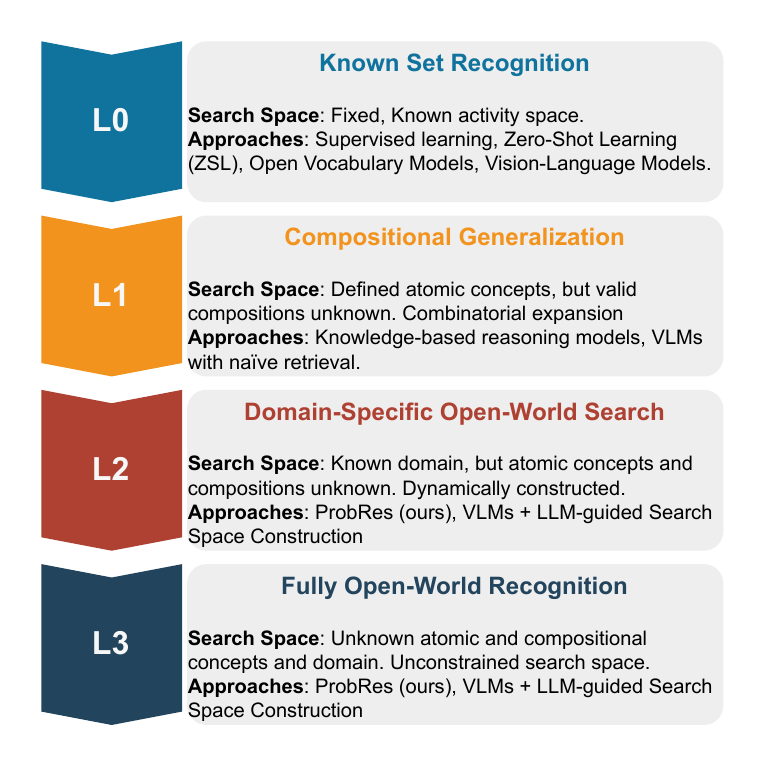}
    \caption{\textit{Taxonomy of Openness in Egocentric Activity Recognition.} We define four levels of openness based on search space constraints: L0 (fixed activity set), L1 (known atomic concepts, unknown compositions), L2 (known domain, inferred activities), and L3 (fully unconstrained search).
    }
    \label{fig:taxonomy}
\end{figure}
Egocentric activity recognition in open-world settings presents a fundamental challenge for intelligent systems, requiring inference from a vast, unconstrained space involving unknown actions and objects, unlike closed-set classification. This adaptive inference is crucial for applications in robotics and AI agents. While Vision-Language Models (VLMs)~\cite{zhao_learning_2022} offer strong zero-shot generalization, their reliance on exhaustive enumeration is inefficient for scalable open-world reasoning. To bridge this gap, structured commonsense knowledge priors and adaptive search strategies are essential. We introduce Probabilistic Residual Search (ProbRes), a novel framework that integrates structured priors with VLM-based likelihood refinement to efficiently navigate unconstrained activity spaces, drastically reducing VLM queries while maintaining or improving recognition accuracy. Our contributions include the ProbRes framework, a hierarchical taxonomy (L0–L3) for open-world challenges, demonstrating ProbRes's efficiency in VLM query reduction, and highlighting VLM text embedding limitations for search effectiveness.

\begin{figure*}
    \begin{tabular}{cc}
    \includegraphics[width=0.55\textwidth]{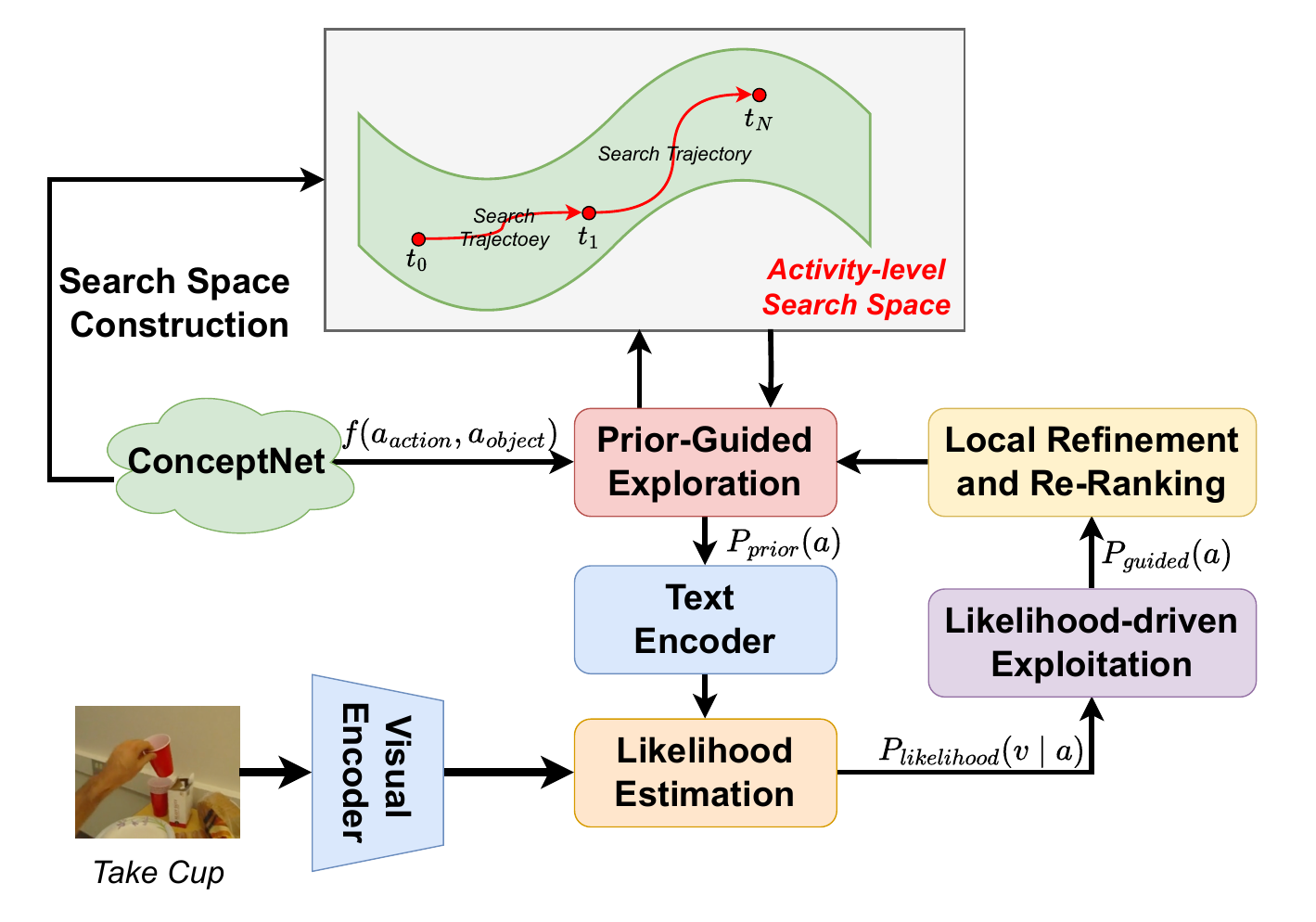} & 
         \includegraphics[width=0.4\textwidth]{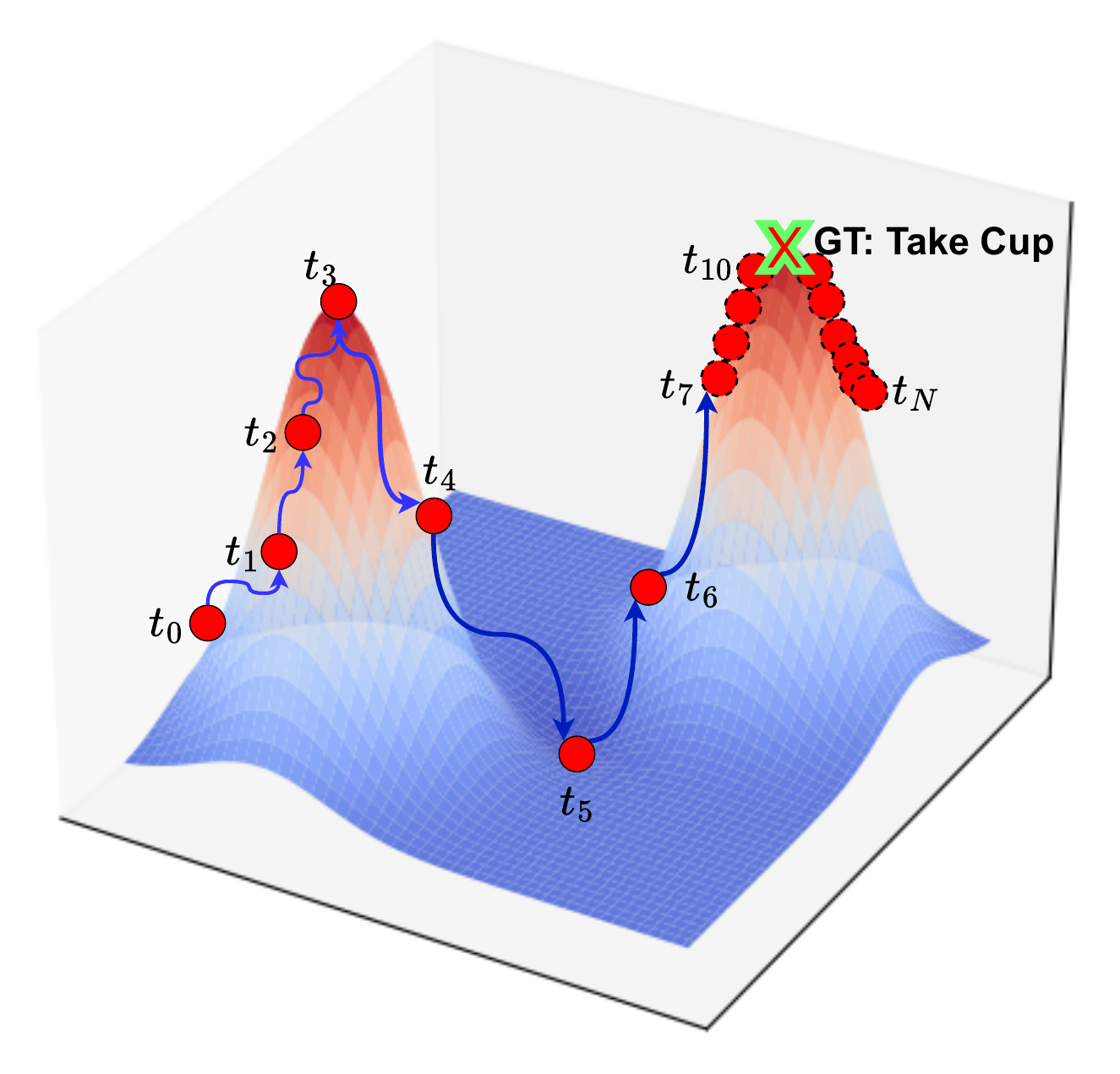} \\
         (a) & (b)\\
    \end{tabular}
    \caption{
    \textbf{(a) ProbRes framework} for open-world egocentric activity recognition. The search space is structured using ConceptNet priors, enabling guided exploration. The model iteratively refines candidates via likelihood estimation, balancing exploration and exploitation, followed by local refinement and re-ranking. (\textbf{b) Search trajectory visualization}, showing how ProbRes navigates likelihood regions to reach high-confidence predictions near the ground truth.
    }
    \label{fig:arch}
\end{figure*}


\section{Related Work}
\textbf{Egocentric video understanding} is a vital area for robotics and AI, with tasks like activity recognition facing challenges, particularly in open-world settings with unconstrained possibilities and unknown entities. While various supervised and self-supervised methods, and more recently multimodal LLM-based models~\cite{lin_egocentric_2022, zhao_learning_2022}, have advanced egocentric action recognition, they often remain data-dependent or struggle with generalization beyond predefined sets. Progress in \textbf{open-world understanding} has been seen in areas like object detection, leveraging \textbf{Vision-Language Models} (VLMs)~\cite{lin_egocentric_2022, zhao_learning_2022} for zero-shot capabilities. However, existing egocentric open-world approaches often rely on limited grounding or iterative refinement, and VLMs themselves, while powerful for zero-shot, are inefficient for large-scale open-world inference due to their reliance on exhaustive enumeration across potential labels. This work introduces ProbRes, a novel framework that addresses these limitations by integrating structured priors and adaptive search mechanisms to efficiently explore and refine activity predictions in unconstrained open-world settings, moving beyond exhaustive VLM inference.

\section{ProbRes: Jump Diffusion-based Reasoning}
\textbf{Overview.} Addressing open-world egocentric activity recognition requires inferring activities from a vast, unconstrained space S, making exhaustive evaluation computationally prohibitive. Given a prior $P_{\text{prior}}(a)$ and VLM likelihood $P_{\text{likelihood}}(v{\mid}a)$, our goal is to efficiently estimate:

\begin{equation}
    a^* = \arg\max_{a \in \mathcal{S}} P_{\text{likelihood}}(v \mid a).
    \label{ideal_label_eqn}
\end{equation}

We propose \textit{Probabilistic Residual Search (ProbRes)}, an adaptive framework guided by symbolic priors from a knowledge graph to navigate the activity text embedding space efficiently. ProbRes operates in three phases: Exploration, Exploitation, and Residual Refinement. 

\subsection{Structured Search Space}We construct a search space combining knowledge-driven priors and VLM embedding organization. A prior probability $P_{\text{prior}}(a)$ over action-object pairs is derived from ConceptNet~\cite{speer_conceptnet_2017} using a semantic affinity score $f(a_{\text{action}}, a_{\text{object}})$, refined by relation-based adjustments. This guides initial search towards semantically plausible activities. The search space is further structured in the VLM text embedding space $\phi(a)$ by sorting activities based on distance to an anchor, ensuring semantic locality for search jumps.
\begin{table*}[ht]
    \centering
    \setlength{\tabcolsep}{2pt} 
    \renewcommand{\arraystretch}{1.2} 
    \resizebox{\textwidth}{!}{%
    \begin{tabular}{|c|cccc|cccc|cccc|cccc|}
        \toprule[1.2pt]
        \vspace{-3pt}
        \multirow{3}{*}{\textbf{Approach}} & \multicolumn{16}{c|}{\textbf{Datasets}} \\
        \cmidrule(lr){2-17}
        & \multicolumn{4}{c|}{\textbf{GTEA Gaze}} & \multicolumn{4}{c|}{\textbf{GTEA Gaze+}} & \multicolumn{4}{c|}{\textbf{EK100}} & \multicolumn{4}{c|}{\textbf{CharadesEgo}} \\
        \cmidrule(lr){2-17}
        & \textbf{VLM Calls} & \textbf{Activity} & \textbf{Phrase} & \textbf{WUPS} & \textbf{VLM Calls} & \textbf{Activity} & \textbf{Phrase} & \textbf{WUPS} & \textbf{VLM Calls} & \textbf{Activity} & \textbf{Phrase} & \textbf{WUPS} & \textbf{VLM Calls} & \textbf{Activity} & \textbf{Phrase} & \textbf{WUPS} \\
        \midrule[1.2pt]
        ALGO & N/A* & 15.06 & 1.80 & 46.89 & N/A* & 18.84 & 3.40 & 43.29 & N/A* & 8.49 & 1.72 & 38.53 & N/A* & 2.62 & 0.10 & 36.68\\
        KGL & N/A** & 6.58 & 0.30 & 35.83 & N/A** & 10.76 & 1.10 & 39.41 & N/A** & 3.07 & 0.94 & 36.85 & N/A** & 2.06 & 0.05 & 31.48\\
        ALGO+LAVILA & N/A* & 22.05 & 3.50 & 49.42 & N/A* & 28.87 & 8.30 & 53.38 & N/A* & 17.69 & 4.39 & 34.47 & N/A* & 2.94 & 0.21 & 37.21\\
        LAVILA-Decomp & 380 & 14.57 & 2.10 & 48.65 & 405 & 26.87 & 7.89 & 53.12 & 29100 & 17.21 & 4.28 & 43.37 & 1254 & 11.82 & 1.51 & 38.99\\
        EGOVLP-Decomp & 380 & 9.31 & 0.90 & 40.51 & 405 & 23.30 & 4.10 & 51.74 & 29100 & 15.14 & 2.05 & 39.94 & 1254 & 10.66 & 1.12 & 38.73\\
        LAVILA & 380 & 29.02 & 9.53 & 51.31 & 405 & 29.82 & 7.91 & 53.27 & 29100 & 18.81 & 4.43 & 43.52 & 1254 & 11.34 & 1.45 & 39.76\\
        EGOVLP & 380 & 17.07 & 1.80 & 46.77 & 405 & 27.00 & 3.95 & 51.48 & 29100 & 12.78 & 2.22 & 39.84 & 1254 & 10.55 & 1.21 & 38.43\\
        \midrule[1.2pt]
        \rowcolor{gray!20} LAVILA + ProbRes (Ours) & 110 & \textbf{33.80} \increase{} & \textbf{8.98} \increase{} & \textbf{53.34} \increase{} & 175 & \textbf{31.70} \increase{} & \textbf{9.33} \increase{} & \textbf{53.82} \increase{} & 3000 & \textbf{18.89} \increase{} & \textbf{4.61} \increase{} & \textbf{43.55} \increase{} & 300 & \textbf{13.71} \increase{} & \textbf{2.12} \increase{} & \textbf{40.91} \increase{}\\
        \rowcolor{gray!20} EGOVLP + ProbRes (Ours) & 110 & \textbf{19.12} \increase{} & \textbf{1.97} \increase{} & \textbf{49.25} \increase{} & 178 & \textbf{29.84} \increase{} & \textbf{6.32} \increase{} & \textbf{53.98} \increase{} & 3000 & \textbf{13.45} \increase{} & \textbf{2.36} \increase{} & \textbf{40.53} \increase{} & 300 & \textbf{12.65} \increase{} & \textbf{1.78} \increase{} & \textbf{38.91} \increase{}\\
        \bottomrule[1.2pt]
    \end{tabular}%
    }
    \caption{
    \textbf{L1 evaluation results} on GTEA Gaze, GTEA Gaze+, EK100, and CharadesEgo. ProbRes consistently improves activity recognition across all datasets while significantly reducing VLM calls, demonstrating its efficiency in structured search. Green arrows indicate performance gains, while red arrows denote declines. *Frame-wise CLIP querying, **Frame-wise Faster R-CNN querying.
    }
    \label{tab:L1_results}
\end{table*}
\begin{table*}[ht]
    \centering
    \setlength{\tabcolsep}{5pt} 
    \renewcommand{\arraystretch}{1.2} 
    \resizebox{\textwidth}{!}{%
    \begin{tabular}{|c|c|c|ccc|ccc|ccc|}
        \toprule
        \multirow{2}{*}{\textbf{Openness}} & \multirow{2}{*}{\textbf{Approach}} & \multirow{2}{*}{\textbf{\# VLM Calls}} & \multicolumn{3}{c|}{\textbf{GTEA Gaze}} & \multicolumn{3}{c|}{\textbf{GTEA Gaze+}} & \multicolumn{3}{c|}{\textbf{EK100}} \\
        \cmidrule(lr){4-12}
        & & & \textbf{Object} & \textbf{Action} & \textbf{Activity} & \textbf{Object} & \textbf{Action} & \textbf{Activity} & \textbf{Object} & \textbf{Action} & \textbf{Activity} \\
        \midrule
        \multirow{4}{*}{\textbf{L2}} 
        & LAVILA & 37191 & {50.72} & 25.96 & 38.34 & 59.25 & \textbf{30.36} & 44.81 & 37.89 & \textbf{23.39} & 30.64 \\
         & \cellcolor{gray!20} LAVILA + ProbRes (Ours) & \cellcolor{gray!20} 1500 & \cellcolor{gray!20} \textbf{53.49}\increase{} & \cellcolor{gray!20} \textbf{33.07}\increase{} & \cellcolor{gray!20} \textbf{43.28}\increase{} & \cellcolor{gray!20} \textbf{62.26}\increase{} & \cellcolor{gray!20} 28.03\decrease{} & \cellcolor{gray!20} \textbf{45.15}\increase{} & \cellcolor{gray!20} \textbf{41.28}\increase{} & \cellcolor{gray!20} 22.56 \decrease{} & \cellcolor{gray!20} \textbf{31.92}\increase{} \\
         \cmidrule(lr){2-12}
        & EGOVLP & 37191 & \textbf{56.77} & 18.34 & 37.56 & 60.77 & 22.27 & 41.52 & 40.87 & \textbf{21.32} & 31.10 \\
        & \cellcolor{gray!20} EGOVLP + ProbRes (Ours) & \cellcolor{gray!20} 1500 & \cellcolor{gray!20} 56.27\decrease{} & \cellcolor{gray!20} \textbf{19.98}\increase{} & \cellcolor{gray!20} \textbf{38.13}\increase{} & \cellcolor{gray!20} \textbf{62.89\increase{}} & \cellcolor{gray!20} \textbf{23.89}\increase{} & \cellcolor{gray!20} \textbf{43.39}\increase{} & \cellcolor{gray!20} \textbf{42.54}\increase{} & \cellcolor{gray!20} 21.07\decrease{} & \cellcolor{gray!20} \textbf{31.81}\increase{} \\
        \midrule
        \multirow{4}{*}{\textbf{L3}} 
        & LAVILA & 195714 & 59.85 & 32.27 & 46.06 & 60.43 & 28.73 & 44.58 & \textbf{41.60} & 20.51 & 31.06\\

        & \cellcolor{gray!20} LAVILA+ProbRes (ours) & \cellcolor{gray!20} 5000 & \cellcolor{gray!20} \textbf{59.89}\increase{} & \cellcolor{gray!20} \textbf{35.53}\increase{} & \cellcolor{gray!20} \textbf{47.71}\increase{} & \cellcolor{gray!20} \textbf{62.35}\increase{} & \cellcolor{gray!20} \textbf{29.04}\increase{} & \cellcolor{gray!20} \textbf{45.70}\increase{} & \cellcolor{gray!20} {41.20}\decrease{} & \cellcolor{gray!20} \textbf{23.96}\increase{} & \cellcolor{gray!20} \textbf{32.58}\increase{}\\
        
        & EGOVLP & 195714 & \textbf{52.94} & 18.93 & 35.94 & 56.95 & 24.37 & 40.66 & \textbf{40.70} & 20.42 & 30.56 \\
        
        & \cellcolor{gray!20} EGOVLP+ProbRes (Ours) & \cellcolor{gray!20} 5000 & \cellcolor{gray!20} {52.01}\decrease{} & \cellcolor{gray!20} \textbf{20.92}\increase{} & \cellcolor{gray!20} \textbf{36.47}\increase{} & \cellcolor{gray!20} \textbf{60.08}\increase{} & \cellcolor{gray!20} \textbf{24.75}\increase{} & \cellcolor{gray!20} \textbf{42.42}\increase{} & \cellcolor{gray!20} {39.91}\decrease{} & \cellcolor{gray!20} \textbf{24.13}\increase{} & \cellcolor{gray!20} \textbf{32.02}\increase{} \\
        \bottomrule
    \end{tabular}%
    }
    \caption{
    \textbf{L2 and L3 evaluation} results on GTEA Gaze, GTEA Gaze+, and EK100. ProbRes achieves significant accuracy gains while reducing VLM calls, demonstrating efficient search in unconstrained spaces. WUPS is reported for object, action, and activity levels.
    }
    \label{tab:L2_L3_results}
\end{table*}
\subsection{Adaptive Search and Refinement}
ProbRes iteratively estimates the most probable activity by balancing exploration (prior-guided sampling using $P_{\text{explore}}(a)$, Eqn ~\ref{eqn:explore_def}) and exploitation (likelihood-guided sampling using $P_{\text{guided}}(a)$, Eqn ~\ref{eqn:exploit_def}). The iterative search optimizes for a score combining search probability and component likelihoods:

\begin{equation}
a^* = \arg\max_{a \in \mathcal{S}} \left[ P_{\text{search}}(a) + \lambda_a S_a + \lambda_o S_o \right]
\label{eq:search_optimization_short}
\end{equation}

where $S_a$ and $S_o$ are action and object component likelihoods.
\begin{equation}
P_{\text{explore}}(a) = \frac{\lambda P_{\text{prior}}(a) + (1-\lambda) \frac{1}{|\mathcal{S}|}}{\sum_{a'} \lambda P_{\text{prior}}(a') + (1-\lambda)}
\label{eqn:explore_def}
\end{equation}

\begin{equation}
P_{\text{guided}}(a) = \frac{P_{\text{prior}}(a) P_{\text{likelihood}}(v \mid a)}{\sum_{a'} P_{\text{prior}}(a') P_{\text{likelihood}}(v \mid a')}
\label{eqn:exploit_def}
\end{equation}

A localized refinement step deterministically re-evaluates top candidates based on VLM likelihoods. Finally, a concept decomposition and re-ranking step computes individual action ($S_a$) and object ($S_o$) alignment scores using $v^T \phi({.})$ to refine the final ranking $S_{\text{final}} = P_{\text{likelihood}}(v \mid a) + \lambda_a S_a + \lambda_o S_o$, mitigating VLM ambiguity and improving robustness.

\textbf{Implementation Details.} We utilize EGOVLP and LAVILA as VLM backbones and ConceptNet for priors. Search spaces for L2/L3 are generated using Gemini 2.0 Flash. Search parameters ($\lambda$, $T$, $\lambda_a$, $\lambda_o$) are tuned for efficiency and accuracy. Average inference time is 2 seconds per video on an RTX 3090.

\section{Experimental Setup}
We evaluate ProbRes on four egocentric activity recognition datasets: GTEA Gaze~\cite{li_learning_2013}, GTEA Gaze+~\cite{fathi_learning_2012}, EPIC-Kitchens-100 (EK100)~\cite{damen_epic-kitchens_2020}, and Charades-Ego~\cite{sigurdsson_actor_2018}, covering both structured and unconstrained egocentric activities. Given the limitations of traditional metrics like accuracy and mAP in open-world settings due to their assumption of a closed label space, we employ WUPS (Wu-Palmer Similarity)~\cite{wu1994verb} to measure semantic similarity and exact phrase-level match accuracy. We compare ProbRes against strong Vision-Language Models (VLMs) such as EGOVLP~\cite{lin_egocentric_2022} and LAVILA~\cite{zhao_learning_2022}, which rely on embedding similarity, and neuro-symbolic approaches like KGL~\cite{aakur_knowledge_2022} and ALGO~\cite{kundu_discovering_2024}, which integrate commonsense reasoning, to demonstrate the advantages of our structured search framework in unconstrained open-world scenarios.

\section{Quantitative Results}\label{sec:results}

\textbf{L1 evaluation} (Table~\ref{tab:L1_results}) shows ProbRes achieves higher accuracy with significantly fewer VLM calls than exhaustive querying. For example, VLM queries are reduced from 380 to 110 on GTEA Gaze and 29,100 to 3,000 on EK100, while improving WUPS and phrase accuracy. This efficiency-accuracy trade-off is vital for scalability, demonstrating effective navigation without brute-force enumeration, even on larger datasets. Neuro-symbolic baselines struggle, highlighting the limitations of fixed knowledge graphs in dynamic environments.

\textbf{L2 Evaluation} (Table~\ref{tab:L2_L3_results}) involves a dynamically constructed, expanded search space with high specificity. ProbRes maintains strong performance and WUPS scores while drastically reducing VLM calls from 37,191 to 1,500. This reinforces that structured priors and guided search enable efficient inference in complex, unconstrained spaces, particularly benefiting action recognition.


\textbf{L3 Evaluation} (Table~\ref{tab:L2_L3_results}) uses a vastly expanded, generic search space across multiple domains, making exhaustive enumeration impractical. ProbRes outperforms baselines despite the less structured space, with the performance gap increasing compared to naive VLM inference. This underscores the critical role of intelligent search strategies and structured priors in highly open-ended environments, suggesting that scalable open-world recognition requires dynamically adaptive search space methods.


\begin{figure}[t]
    \centering
    \includegraphics[width=0.99\linewidth]{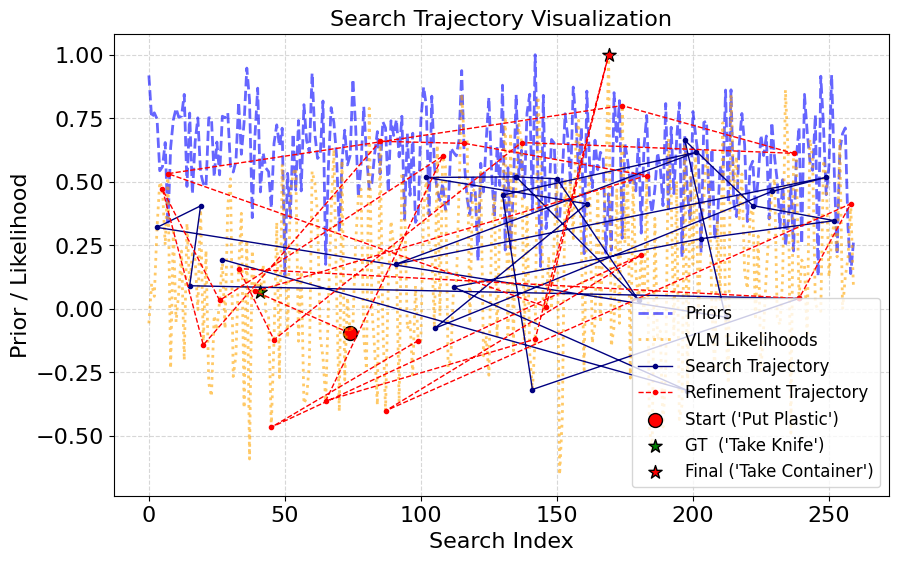}
    \caption{\textbf{Qualitative Visualization} of the search trajectory by ProbRes across different phases indicating exploration,  exploitation, and the final refinement phase.}
    \label{fig:Trejectory_Vis}
\end{figure}

\textbf{Qualitative Analysis} (Figure~\ref{fig:Trejectory_Vis}) visualizes the search trajectory. Exploration samples diverse candidates guided by priors, while exploitation narrows to plausible ones using likelihood. Refinement further filters semantically similar incorrect activities. Inconsistencies in VLM embedding distances highlight the need for more structured representations to improve search efficiency in open-world settings.

\section{Limitations, Discussion, and Future Work}
While ProbRes advances open-world egocentric activity recognition by replacing brute-force enumeration with structured search, challenges remain. Reliance on VLM biases and unstructured text embeddings can lead to inefficient search trajectories, suggesting the need for better semantic organization. LLM-generated search spaces, while enabling scalability, require careful curation. Further efficiency improvements are necessary for real-world deployment. ProbRes demonstrates the potential of integrating commonsense priors with probabilistic search for scalable and interpretable egocentric AI. Future work will explore incorporating additional visual cues, like human-object interactions, to enhance inference.

\textbf{Acknowledgement.} This research was supported in part by the US NSF grants IIS 2348689 and IIS 2348690.

{
    \small
    \bibliographystyle{ieeenat_fullname}
    \bibliography{main}

\begin{thebibliography}{10}
\providecommand{\natexlab}[1]{#1}
\providecommand{\url}[1]{\texttt{#1}}
\expandafter\ifx\csname urlstyle\endcsname\relax
  \providecommand{\doi}[1]{doi: #1}\else
  \providecommand{\doi}{doi: \begingroup \urlstyle{rm}\Url}\fi

\bibitem[Aakur et~al.(2022)Aakur, Kundu, and Gunti]{aakur_knowledge_2022}
Sathyanarayanan~N. Aakur, Sanjoy Kundu, and Nikhil Gunti.
\newblock Knowledge guided learning: {Open} world egocentric action recognition with zero supervision.
\newblock \emph{Pattern Recognition Letters}, 156:\penalty0 38--45, 2022.
\newblock Publisher: North-Holland.

\bibitem[Damen et~al.(2020)Damen, Doughty, Farinella, Fidler, Furnari, Kazakos, Moltisanti, Munro, Perrett, Price, and Wray]{damen_epic-kitchens_2020}
Dima Damen, Hazel Doughty, Giovanni~Maria Farinella, Sanja Fidler, Antonino Furnari, Evangelos Kazakos, Davide Moltisanti, Jonathan Munro, Toby Perrett, Will Price, and Michael Wray.
\newblock The {EPIC}-{KITCHENS} {Dataset}: {Collection}, {Challenges} and {Baselines}, 2020.
\newblock arXiv:2005.00343 [cs].

\bibitem[Fathi et~al.(2012)Fathi, Li, and Rehg]{fathi_learning_2012}
Alireza Fathi, Yin Li, and James~M. Rehg.
\newblock Learning to {Recognize} {Daily} {Actions} {Using} {Gaze}.
\newblock In \emph{Computer {Vision} – {ECCV} 2012}, pages 314--327, Berlin, Heidelberg, 2012. Springer.

\bibitem[Kundu et~al.(2024)Kundu, Trehan, and Aakur]{kundu_discovering_2024}
Sanjoy Kundu, Shubham Trehan, and Sathyanarayanan~N. Aakur.
\newblock Discovering {Novel} {Actions} from {Open} {World} {Egocentric} {Videos} with {Object}-{Grounded} {Visual} {Commonsense} {Reasoning}, 2024.
\newblock arXiv:2305.16602 [cs].

\bibitem[Li et~al.(2013)Li, Fathi, and Rehg]{li_learning_2013}
Yin Li, Alireza Fathi, and James~M. Rehg.
\newblock Learning to {Predict} {Gaze} in {Egocentric} {Video}.
\newblock In \emph{2013 {IEEE} {International} {Conference} on {Computer} {Vision}}, pages 3216--3223, Sydney, Australia, 2013. IEEE.

\bibitem[Lin et~al.(2022)Lin, Wang, Soldan, Wray, Yan, Xu, Gao, Tu, Zhao, Kong, Cai, Wang, Damen, Ghanem, Liu, and Shou]{lin_egocentric_2022}
Kevin~Qinghong Lin, Alex~Jinpeng Wang, Mattia Soldan, Michael Wray, Rui Yan, Eric~Zhongcong Xu, Difei Gao, Rongcheng Tu, Wenzhe Zhao, Weijie Kong, Chengfei Cai, Hongfa Wang, Dima Damen, Bernard Ghanem, Wei Liu, and Mike~Zheng Shou.
\newblock Egocentric {Video}-{Language} {Pretraining}, 2022.
\newblock arXiv:2206.01670 [cs].

\bibitem[Sigurdsson et~al.(2018)Sigurdsson, Gupta, Schmid, Farhadi, and Alahari]{sigurdsson_actor_2018}
Gunnar~A. Sigurdsson, Abhinav Gupta, Cordelia Schmid, Ali Farhadi, and Karteek Alahari.
\newblock Actor and {Observer}: {Joint} {Modeling} of {First} and {Third}-{Person} {Videos}, 2018.
\newblock arXiv:1804.09627 [cs].

\bibitem[Speer et~al.(2017)Speer, Chin, and Havasi]{speer_conceptnet_2017}
Robyn Speer, Joshua Chin, and Catherine Havasi.
\newblock {ConceptNet} 5.5: {An} {Open} {Multilingual} {Graph} of {General} {Knowledge}.
\newblock \emph{Proceedings of the AAAI Conference on Artificial Intelligence}, 31\penalty0 (1), 2017.
\newblock Number: 1.

\bibitem[Wu and Palmer(1994)]{wu1994verb}
Zhibiao Wu and Martha Palmer.
\newblock Verb semantics and lexical selection.
\newblock In \emph{Proceedings of the 32nd annual meeting on Association for Computational Linguistics}, pages 133--138, 1994.

\bibitem[Zhao et~al.(2022)Zhao, Misra, Krähenbühl, and Girdhar]{zhao_learning_2022}
Yue Zhao, Ishan Misra, Philipp Krähenbühl, and Rohit Girdhar.
\newblock Learning {Video} {Representations} from {Large} {Language} {Models}, 2022.
\newblock arXiv:2212.04501 [cs].

\end{thebibliography}
}


\end{document}